\documentclass{article}

\usepackage[final]{neurips_2026}
\makeatletter
\renewcommand{\@noticestring}{Workshop on Who Verifies the Agents? Toward Reliable Agent Development, NeurIPS 2026.}
\makeatother

\usepackage[utf8]{inputenc} % allow utf-8 input
\usepackage[T1]{fontenc}    % use 8-bit T1 fonts
\usepackage{hyperref}       % hyperlinks
\usepackage{url}            % simple URL typesetting
\usepackage{booktabs}       % professional-quality tables
\usepackage{amsfonts}       % blackboard math symbols
\usepackage{amsmath}        % display math (patched for lineno by neurips_2026)
\usepackage{array}          % for column formatting in tables
\usepackage{nicefrac}       % compact symbols for 1/2, etc.
\usepackage{microtype}      % microtypography
\usepackage{xcolor}         % colors

\usepackage{graphicx}       % \includegraphics and \resizebox
\usepackage{enumitem}       % enumerate [leftmargin=*]
\usepackage{listings}       % lstlisting code blocks
\lstdefinestyle{plain}{basicstyle=\small\ttfamily,breaklines=true}
\lstdefinestyle{prompt}{
basicstyle=\small\ttfamily,
breaklines=true,
breakatwhitespace=false,
columns=fullflexible,
keepspaces=true,
backgroundcolor=\color{gray!8},
frame=single,
framerule=0.4pt,
rulecolor=\color{gray!50},
xleftmargin=4pt,
xrightmargin=4pt,
aboveskip=4pt,
belowskip=4pt,
}
\usepackage{tikz}           % pipeline figure
\usetikzlibrary{arrows.meta,positioning}

\definecolor{softblue}{RGB}{204,229,255}
\definecolor{warnyellow}{RGB}{255,242,204}
\definecolor{okgreen}{RGB}{212,239,223}
\definecolor{badred}{RGB}{248,215,218}

\title{When Terminal-Agent Training Stalls: Demystifying Data Generation and Verification Challenges}
\author{
  Xi Qin \\
  SAP Lab \\
  Palo Alto, CA 94304 \\
  \texttt{grace.qin@sap.com} \\
  \And
  Isabel Kurth\thanks{This work was done when Isabel was an intern at SAP.} \\
  SAP Lab \\
  Palo Alto, CA 94304 \\
  \texttt{isabel.kurth@t-online.de} \\
  \And
  Xin Cui \\
  SAP Lab \\
  Palo Alto, CA 94304 \\
  \texttt{xin.cui@sap.com} \\
  \AND
  Elin Park \\
  SAP Lab \\
  Palo Alto, CA 94304 \\
  \texttt{elin.park@sap.com} \\
  \And
  Alexander Schaefer \\
  SAP Lab \\
  Palo Alto, CA 94304 \\
  \texttt{alexander.schaefer01@sap.com} \\
  \And
  Yaad Oren \\
  SAP Lab \\
  Palo Alto, CA 94304 \\
  \texttt{yaad.oren@sap.com} \\
}

\begin{document}

\maketitle

\begin{abstract}
% ==========================================================================

%Training terminal agents with reinforcement learning requires environments that are both executable and learnable: tasks must be solvable by the policy being trained, not merely structurally valid. We extend Endless Terminals~\cite{gandhi2025endless}, a pipeline that procedurally generates terminal-agent tasks, along three dimensions. First, we migrate the runtime from Apptainer to Docker and adopt the Harbor benchmark API~\cite{harbor}, gaining pre-agent validation, parallel execution, and a standardized evaluation harness. Second, we redesign the task-generation prompt around engineer voice, anti-leakage, and domain coverage, adding three-level cognitive difficulty and pre-agent validation that rejects structurally invalid tasks before rollout. Third, we generate a new dataset with 8192-token context windows, raising solvability from approximately 10\% to 56.5\% ($\mathrm{pass}@1$ under Claude~4.6 Sonnet) across 3{,}310 tasks. Training experiments show that calibration, not structural validity alone, controls the available learning signal. Qwen2.5-3B PPO on the original sparse dataset fails to improve beyond its 7.8\% base $\mathrm{pass}@1$; for Qwen3.5-9B GRPO, adding hard tasks reduces mean rollout $\mathrm{pass}@2$ from 81.3\% to 20.6\% under an otherwise fixed configuration, restoring failure cases for learning.

%%%%%%%%%%%

Automated terminal-agent training relies on generated environments and verifiers to turn open-ended interaction into reward. Yet a runnable Docker image and executable test suite do not guarantee that the resulting reward is faithful, informative, or even attributable to agent behavior. We present a pipeline that generates containerized terminal tasks and their executable verifiers, and use it to demystify failure modes in task generation and live-environment evaluation. Our analysis separates four requirements that are often conflated: structural validity, verifier faithfulness, policy solvability, and infrastructure reliability. In a Qwen3.5-9B GRPO training, the model overfits on a relatively easy dataset in 20 steps. In contrast, in a controlled Qwen3.5-9B comparison, adding hard tasks reduces mean rollout $\mathrm{pass}@2$ from 81.3\% to 20.6\%, restoring informative failures without changing the training configuration. We further identify duplicate task generation, incomplete anti-shortcut safeguards, Dockerfile failures, asynchronous $\mathrm{pass}@k$ bias, and environment errors caused by concurrency constraints. All impact the precision in measurement. These findings should motivate the community for a terminal agent specific evaluation harness that pairs structural and semantic checks, and task solvability with extensive verifier audits.
\end{abstract}

% ==========================================================================

\section{Introduction}
% ==========================================================================

% Stakes: terminal agents and the RL environment problem.
Autonomous terminal agents, as models that issue shell commands, read outputs,
and iterate to complete engineering tasks, are a promising direction for
automating software operations.
Reinforcement learning is a natural training approach: either a binary pass-or-fail signal or multi-state checks from
a test suite, provides a clean reward without human labeling.
The bottleneck lies in environment harness.
Writing executable terminal tasks with correct initial states, verifiable
final states, and graded rewards is labor-intensive, and the resulting
datasets are small.

% Problem gap: structural validity is not the same as learnability.
Gandhi et al.~\cite{gandhi2025endless} address environment construction with
Endless Terminals, a pipeline that procedurally generates tasks without human
annotation.
Their pipeline produces structurally valid tasks: the container boots, the test suite runs, and this process discards roughly half of all generated candidates by OpenAI O3 model, one of the top models at that time.
This seemingly sparse reward would highly possibly make RL training difficult: even with 16 samples per task, most
batches return all-zero reward and produce no gradient. Future work is left to quantify the impacts of this sparsity issue on training instability.

In this work, we aim to bridge this gap by first analyzing the challenges of data generation and verification in this brittle pipeline.

\textbf{Endless Terminals and Harbor}

Endless Terminals~\cite{gandhi2025endless} generates terminal-agent tasks
in five stages: task description and ground truth, initial-state pytest,
final-state pytest, Apptainer container definition, and packaged task bundle.
Each stage is a batched LLM call conditioned on earlier outputs.
The pipeline requires no human annotation and scales with inference computation.
Structure-wise, we preserve this five-stage skeleton and replace the pipelines of data generation and curation, evaluation harness, and LLM
providers.

Harbor~\cite{harbor} is an open-source benchmark framework for terminal agents.
It prescribes a fixed directory layout per task: an instruction file, a
\texttt{task.toml} metadata file, a Docker environment, a reference solution,
and a pytest verifier that writes binary reward.
Harbor's CLI builds task images, attaches an agent, drives terminal interaction,
and collects results.
We adopt Harbor as our task format and evaluation harness.

%\subsection{Reward Sparsity in Terminal RL}

%Prior work on RL for code generation~\cite{deepseekcoder,swebench} shows that sparse binary reward can drive learning when the base model already solves some fraction of training tasks. The critical quantity is the per-batch pass rate: if most batches return all-zero reward, the policy gradient estimate has high variance and zero mean, and training stalls. Terminal tasks are harder than code-generation tasks because the action space is broader (any shell command touching all layers of a computer system) and the environment is more complex with propagating intermediate errors and noisy evaluation signals. 

% Contributions.
We make three contributions.
\begin{enumerate}[leftmargin=*]
    %\item \textbf{Migration to Docker and Harbor.} We adapt the Endless Terminals pipeline to emit Harbor-compatible task directories with Dockerfiles, pre-agent validation, and reference    solution generation. Pre-agent validation rejects tasks where the initial-state test fails or the final-state test already passes -- catching container/test mismatches    before rollout.

    \item \textbf{Terminal Task Dataset}
    We generate 18{,}185 tasks with 8192-token context windows from Claude Opus models, \textbf{the largest} to our best knowledge.

    \item \textbf{Task Quality Improvements.}
    We design the generation prompts thoroughly around three dimensions: engineer voice
    (casual, storytelling with deliberate imperfections), anti-leakage
    (prohibiting restatement of verification criteria), and domain coverage
    (round-robin balancing across eight task categories).
    We replace syntactic complexity levels with three levels of  cognitive difficulty
    and thread difficulty through every generation stage.

    \item \textbf{Structural Validity Is Not Benchmark Validity.} We demystify three failure modes in automated terminal task generation and verification: task calibration, verifier loopholes, and live-environment
infrastructure failures can corrupt the reward signal used to evaluate and train agents.

\end{enumerate}

% Results preview.
%On the original dataset (457 tasks, $\sim$10\% solvable), Qwen2.5-3B PPO runs 228 steps without improvement: the base model's 7.8\% $\mathrm{pass}@1$ is the best checkpoint. For Qwen3.5-9B GRPO, adding a hard-task subset to an otherwise fixed training configuration reduces mean rollout $\mathrm{pass}@2$ from 81.3\% to 20.6\%. Together, these results show that RL training on terminal tasks requires careful, model-specific dataset curation.

\section{Acknowledgment}

The authors gratefully acknowledge Kanishk Gandhi of Stanford University Human-Centered Artificial Intelligence Institution, for his participation in constructive discussions that helped sharpen the analysis and presentation of this work.

\section{Data Generation}
% ==========================================================================

\begin{figure}[t]
\centering
\resizebox{\textwidth}{!}{%
\begin{tikzpicture}[
    node distance=1.1cm and 1.25cm,
    box/.style={rectangle, rounded corners, draw=black!55, thick, align=center,
                minimum width=3.0cm, minimum height=1.0cm, font=\small},
    gen/.style={box, fill=softblue},
    check/.style={box, fill=warnyellow},
    ok/.style={box, fill=okgreen},
    bad/.style={box, fill=badred},
    arrow/.style={-{Latex[length=2.4mm]}, thick, draw=black!70},
    note/.style={rectangle, draw=black!40, rounded corners, align=left,
                 font=\scriptsize, fill=white, text width=4.1cm}
]

\node[gen] (template) {1. Task template\\instruction + truth};
\node[gen, right=of template] (initial) {2. Initial state\\pytest};
\node[gen, right=of initial] (final) {3. Final state\\pytest};
\node[gen, right=of final] (docker) {4. Dockerfile\\environment};
\node[gen, right=of docker] (save) {5. Harbor task\\directory};

\draw[arrow] (template) -- (initial);
\draw[arrow] (initial) -- (final);
\draw[arrow] (final) -- (docker);
\draw[arrow] (docker) -- (save);

\node[check, below=of docker] (pre) {Pre-agent validation\\initial passes; final fails};
\draw[arrow] (docker) -- (pre);
\draw[arrow] (pre.east) -- ++(0.5,0) |- (save.south);

\node[ok, below=of save] (run) {Solution sampling\\N attempts per task};
\draw[arrow] (save) -- (run);

\node[ok, below left=of run] (solved) {Solvable\\$\geq 1$ passing attempt};
\node[bad, below right=of run] (unsolved) {Unsolved\\0 passing attempts};
\draw[arrow] (run) -- (solved);
\draw[arrow] (run) -- (unsolved);

\node[note, below=0.55cm of solved] (solvednote) {Use for RL training, reference solutions, and curriculum construction.};
\node[note, below=0.55cm of unsolved] (unsolvednote) {Triage: genuinely hard, ambiguous, impossible environment, or brittle test.};
\draw[arrow] (solved) -- (solvednote);
\draw[arrow] (unsolved) -- (unsolvednote);

\end{tikzpicture}%
}
\caption{Overall pipeline. Stages 1--5 generate Harbor
task directories. Pre-agent validation filters structural failures before rollout.
Solution sampling assigns empirical solvability labels, which control RL training
signal.}
\label{fig:pipeline}
\end{figure}
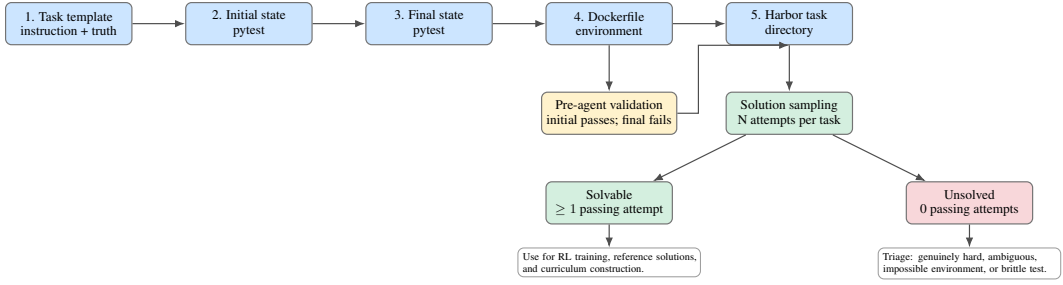

In our five-stage pipeline (Figure \ref{fig:pipeline}), stages 1--3 produce the task description, ground truth, initial-state tests, and final-state tests.
Stage 4 produces a Dockerfile to build the sandbox for simulation and verification of a terminal task. Stage 5 writes a Harbor task directory instead of a flat bundle. There are three types of models involved across stages: 1. Generation model for task generation; 2. Generation model for solution generation; 3. The agent policy model. We use Claude Opus 4.6/4.8 for task generation. Following a distillation-inspired approach, we use a less capable model for solution generation than for task generation. For example, when Claude 4.8 Opus generates tasks, Claude 4.8 Sonnet generates these tasks' solutions. The agent policy model is then limited to smaller parameter scales (3B, 4B, or 9B) to reduce training and inference costs.

Here are the details in each stage.

\subsection{From Apptainer to Docker and Harbor}

The original pipeline targets Apptainer because SLURM clusters require
daemonless unprivileged containers.
Our setting infuses Harbor, which expects a Dockerfile per task.
Stage 4's prompt was rewritten around specific constraints in Docker: heredoc
handling, the absence of \texttt{systemd}, restrictions on Docker-in-Docker,
non-root user execution, and wrapper \texttt{CMD} scripts for tasks that need
a service running at agent connect time. The Harbor task directory layout is listed in Appendix \ref{appendix:harbor}. During training and evaluation, the verifier writes reward $1$ when the final-state pytest passes and $0$
otherwise.
We follow this convention throughout.

\subsection{Task-Template Prompt Redesign}

The original prompt is approximately thirty lines and specifies a rubric for
the \texttt{<task>} and \texttt{<truth>} XML blocks.
It produces two systematic failure modes.

\textit{Worksheet Voice.} The model produces tasks that sound like graded assignments: numbered requirements, phrases such as ``The output
should\ldots'' and ``Ensure that\ldots,'' and openings like ``Please complete
the following task.''
These patterns signal to the agent that it is being evaluated, not helping.

\textit{Grader Leakage.} The model restates verification criteria inside the
task body. For example, asking the agent to ``produce a log of every change''
when the verifier checks exactly that log.
Leakage collapses the task from problem solving to transcription.

We redesign the prompt around three dimensions.

\textbf{Voice.} The narrator is an engineer: casual, and whom could be possibly wrong.
The prompt enumerates opening modes -- trigger/debug, build, reverse
engineering, recovery, migration, investigation, optimization -- and requires
the model to rotate across them.
Each task must contain at least one deliberate imperfection: a hedge, a
possibly wrong guess, or a register cue.
Banned phrasings are listed verbatim so the model can self-check before
emitting.

\textbf{Anti-leakage.} The prompt enumerates leakage behaviors: restating
verification criteria, requesting artifacts whose only purpose is grading,
enumerating sub-requirements, and explaining the root cause in the task body.
The \texttt{<truth>} section requires four items in order: initial state,
expected final state, invariants, and anti-shortcut guards.

\textbf{Domain Coverage.} To prevent the model from collapsing to a small
attractor (in our case, ``Python data bug''), the system message lists
multiple domains and task types and instructs deliberate rotation.
A self-check asks three questions: did the diagnosis leak, does the task
contain at least one imperfection, and does the first sentence drop the agent
into a situation rather than announcing a project.

The system message carries standing rules.
The user message offers task variation: a domain drawn from balanced
category buckets, a fictitious narrator persona, and a difficulty level with
level-specific guidance. We share the core prompts in Appendix \ref{appendix:prompts}.

\subsection{Difficulty and Category Balancing}

Previous work in \cite{gandhi2025endless} defines six complexity levels by command count, sampled
uniformly and inserted as a bare phrase.
We think command count is a poor proxy for solve difficulty: \texttt{chmod +x foo} and
a one-line \texttt{sed} bug fix are both one-command tasks. So we replace it with a three-level \texttt{difficulty} dimension defined in
cognitive terms.
Easy tasks have a single obvious cause and a direct fix.
Medium tasks require investigation before action.
Hard tasks satisfy at least one of five criteria: misleading symptom,
conflicting constraints, required domain knowledge, large state space, or
reverse engineering.
Each level includes examples rather than a bare phrase.
Difficulty is threaded through every generation stage and written into
\texttt{task.toml}, aligning with Harbor's metadata schema.

The core round-robin sampling strategy applies across ten high-level buckets enforces category balance:
\texttt{file\_ops}, \texttt{networking}, \texttt{services},
\texttt{security}, \texttt{debugging}, \texttt{data\_processing},
\texttt{git\_ops}, \texttt{system\_admin}, \texttt{build\_tools}, and
\texttt{text\_processing}.
Without this, a single batched call produces correlated tasks: if the first
task involves \texttt{git rebase}, several subsequent tasks do as well.

\subsection{Pre-Agent Validation}

After Stage 4 builds the Docker image, the pipeline runs the generated tests
against the container before any agent rollout.
A task survives only when the initial-state test passes and the final-state
test fails. The initial pass requirement checks that the Dockerfile correctly seeds the
world described in the ground truth. Missing input files, wrong port bindings, or incorrect permissions are caught here rather than during rollout.

%The final-fail requirement rejects tasks already solved at time zero. If the final-state condition holds in the initial container, Harbor returns reward $1$ regardless of agent behavior, making the task useless as training signal.

This pre-agent validation step is absent from the pipeline in \cite{gandhi2025endless},
which discovers such mismatches only during solution sampling.

\subsection{Solution Generation}

After task generation, a solution generation stage samples agent trajectories
using Harbor's \texttt{Terminus2} agent.
The companion script \texttt{collect\_harbor\_results.py} scans Harbor job
directories, groups trials by task, and writes \texttt{solution.json} with
per task $\mathrm{pass}@k$ metrics.

Solution generation assigns an empirical solvability label to each task.
Tasks with at least one passing attempt are candidates for RL training
datasets.
Tasks with zero passing attempts require triage: they may be genuinely hard,
underspecified, impossible, or over strictly graded.

%\subsection{Pipeline Summary}

%Figure~\ref{fig:pipeline} shows the full pipeline. The key structural addition relative to Gandhi et al.\ is the bifurcation between \emph{structural validity} (initial-state pass, final-state fail) and \emph{empirical solvability} (at least one passing agent trajectory). Structural validity is necessary but not sufficient for RL training; empirical solvability is the controlling variable.

\section{Dataset Analysis}
% ==========================================================================

\subsection{Solvability of Generated Tasks}

We generate three datasets as summarized in Table~\ref{tab:datasets}. Details of assigning categories are in Appendix \ref{appendix:category-strategy}. We identify 8 attempts being a sweet point for most solution generation, which is proved by figure~\ref{fig:passk} in Appendix.

%\textbf{$V1$} uses the original 4096-token context window and the original generation prompt; it yields 457 training tasks with approximately 10\% $\mathrm{pass}@1$ under Claude~4.6 Sonnet. \textbf{$V2$} uses 8192-token context windows and the redesigned prompt;
%it yields 3{,}310 tasks across two parallel generation runs, with 56.5\% $\mathrm{pass}@1$ (2{,}092 / 3{,}310 tasks solvable) under Claude~4.6 Sonnet After solvability filtering, $V2$ provides 1{,}882 training tasks and 210 validation tasks.

The $5.6\times$ improvement in solvability from $V1$ to $V2$ comes from two sources.
Longer context allows more complete environment specifications, reducing cases where the agent cannot infer the missing state.
The redesigned prompt reduces grader leakage and worksheet voice, producing
tasks that an agent can approach as genuine engineering problems.

\begin{table}[h]
\centering
\caption{Dataset overview. For solvability measured by $\mathrm{pass}@1$, Claude~4.6 Sonnet filters for $V1$ and $V2$, and Claude~4.7 Opus for $V3$ specifically.}
\label{tab:datasets}
\begin{tabular}{lccccc}
\toprule
\textbf{Dataset} & \textbf{Model} & \textbf{Context} & \textbf{Total tasks} &
    \textbf{$\mathrm{pass}@1$} & \textbf{Category Assignment} \\
\midrule
$V1$ (Dataset-4k) & Opus 4.6 & 4096 & 5038 & $\sim$10\% & Round-robin \\
$V2$ (Dataset-8k) & Opus 4.6 & 8192 & 4929 & 56.5\% & Difficulty and category coupling \\
$V3$ (Dataset-5k) & Opus 4.8 & 8192 & 8218 & 39.6\% & Difficulty first, category next \\ 
\bottomrule
\end{tabular}
\end{table}

\subsection{Difficulty and Category Breakdown}
% Key abstraction: the solvability band.

%Raising solvability is necessary but not sufficient, the model must still fail on some tasks at training time to learn from contrastive examples.
Overall, to avoid heavy tails in easy and hard, we make medium tasks 63\%, then easy and hard 18.5\% separately. 
Within $V3$, difficulty is distributed as follows:
easy (57\%), medium (36\%), and hard (7\%), after solvability filtering.
%Figure~\ref{fig:difficulty} shows that 
A subset of randomly sampled 1{,}000 tasks, has pass rate that monotonically decreases
with difficulty (easy: 97\%, medium: 92\%, hard: 89\%). This confirms the
cognitive difficulty definitions are meaningful.

Round-robin construction makes category coverage near uniform.
The eight common task domains are shown in Figure~\ref{fig:category_desc} left graph; domain distribution remains in a universal pattern across all datasets.

%\begin{figure}[t]
%\centering
%\includegraphics[width=\columnwidth]{figures/difficulty_breakdown.pdf}
%\caption{Left: task counts by difficulty level in $V2$ (after solvability
%filtering). Right: mean $\mathrm{pass}@1$ by difficulty (Claude~4.6 Sonnet); easy tasks
%are solved 97\% of the time, hard tasks 89\%.}
%\label{fig:difficulty}
%\end{figure}

\subsection{Instruction Length Distribution}

\begin{figure}[t]
\centering
\includegraphics[width=0.49\columnwidth]{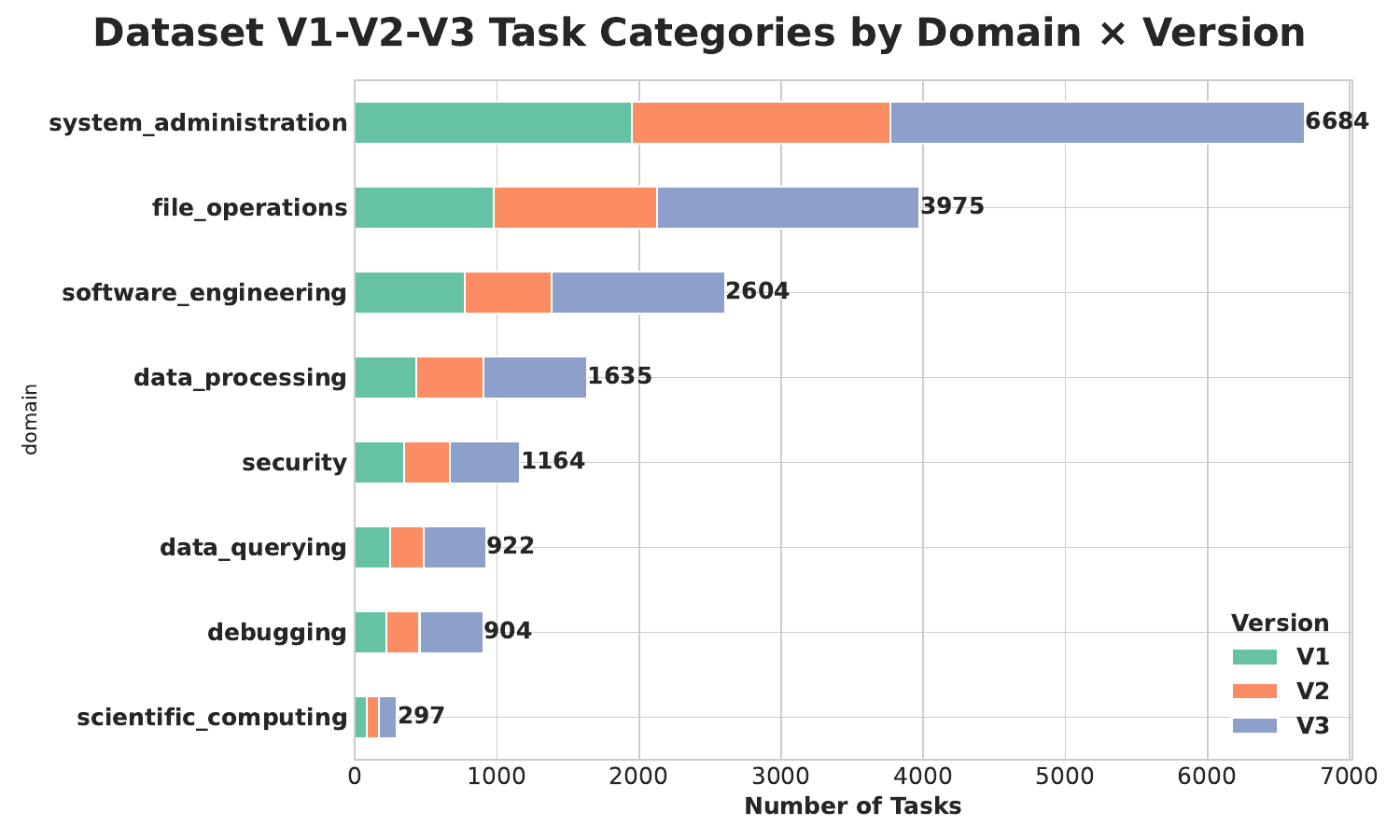}
\hfill
\includegraphics[width=0.49\columnwidth]{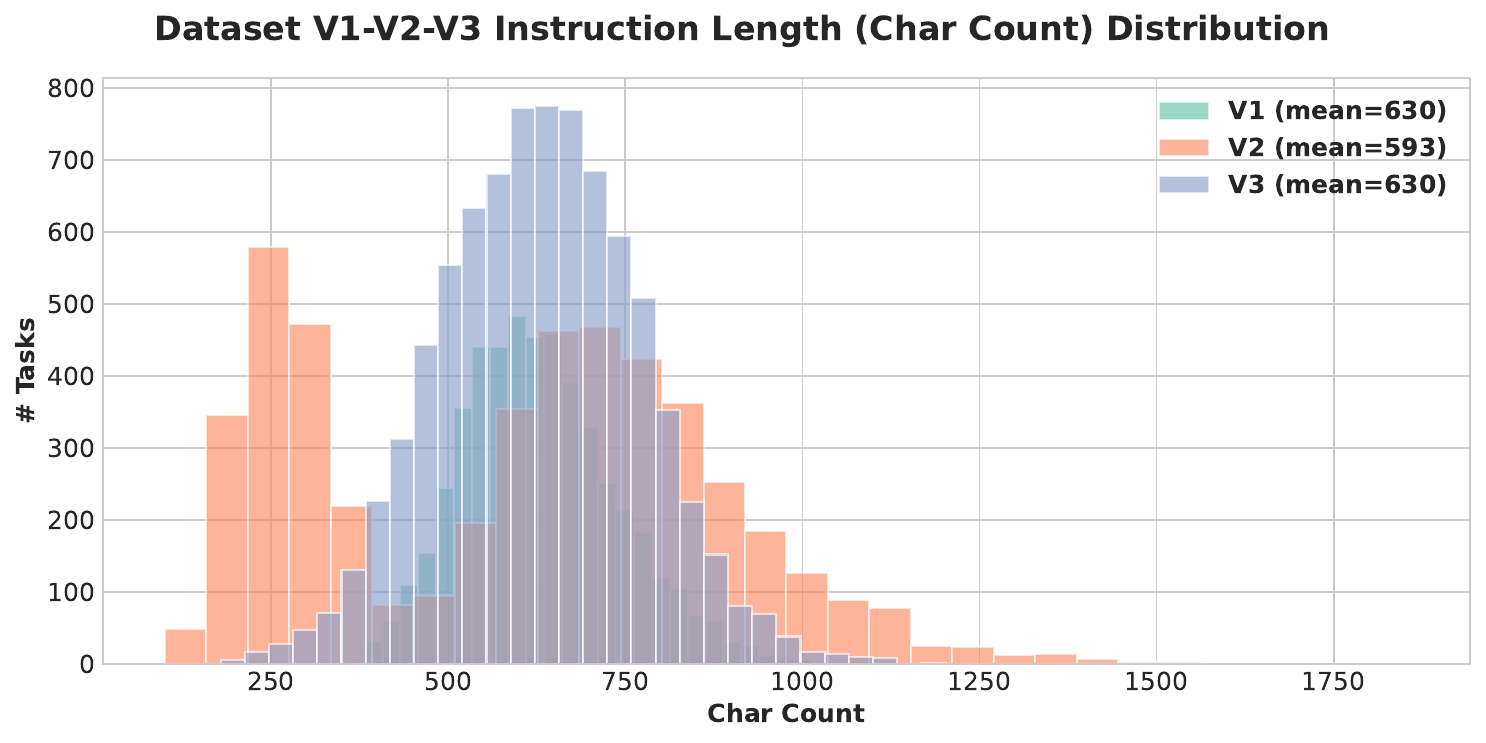}
\caption{Left: task categories in all three datasets; no single category
dominates.
Right: instruction length distributions for all datasets.}
\label{fig:category_desc}
\end{figure}

In Figure \ref{fig:category_desc} right graph, $V2$ instructions has a broader coverage than $V1$ and $V3$, with some extremely long instructions with 1000 more characters, reflecting richer environment specifications
enabled by the larger context window. But the caveat is the heavy left tail down to 200 characters, meaning a significant number of unusually short instructions that could detail the trajectory. $V3$ covers a broader range than $V1$ with a nice bell-shaped and tightly centered distribution, which shows a more balanced instruction design. 

\subsection{Verifier Quality}

Assertion count improved steadily from $V1$ (mean=24.3) to $V3$ (mean=27.6), shifting noticeably rightward and having the least low-count tail, suggesting more thorough verifiers over time. However, the semantic type breakdown reveals a trade-off: $V1$ had the highest process output assertions (5.7 per task), which degraded to 2.7 in $V3$. File content assertions similarly collapsed from 5.8 in $V2$ to 1.7 in $V3$, as shown in \ref{fig:assertion_semantic_types}. We show more details in graphs in Appendix \ref{appendix:data-quality}.

\begin{figure}[h]
\centering
\includegraphics[width=\columnwidth]{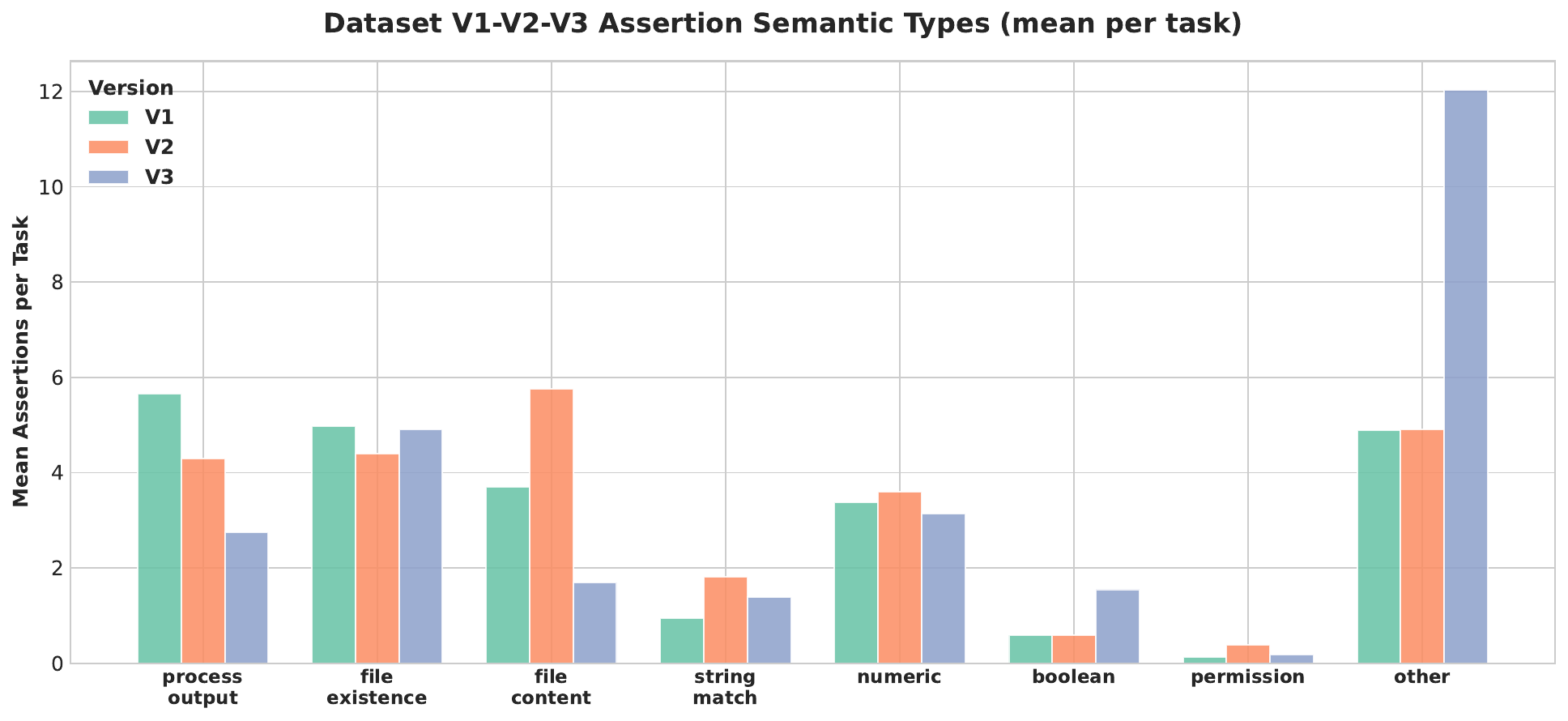}
\caption{Assertion semantic types across $V1$, $V2$, and $V3$ datasets.}
\label{fig:assertion_semantic_types}
\end{figure}
% ==========================================================================
\section{Training Experiments}
% ==========================================================================

We use training rollouts to test whether a generated task distribution supplies
useful RL signal, rather than treating a runnable verifier as sufficient.
All experiments use SkyRL~\cite{skyrl1}\cite{skyrl2}, binary reward from Harbor's verifier,
and the \texttt{Terminus2} agent with at most eight turns per rollout.

Before training started, we should assess the solvability baseline, i.e. the range of per-task pass rates where RL training can make progress. In extreme cases,
too low rate (near 0\%) causes the absence of gradient signals; too high rate (near 100\%) causes marginal training gains. Table \ref{tab:experiments} shares these baselines. As a ceiling reference, Claude Opus 4.6's solve rate ($\mathrm{pass}@1$) over three validation datasets (100, 151, 517 samples respectively from $V1$ to $V3$) from $E1$ to $E3$ are 94.1\%, 98.0\% and 11.2\% respectively; similarly, Qwen3.5-9B has the solve rate in the same three datasets at 86.3\%, 92.0\%, 9.8\% respectively.

\begin{table}[h]
\centering
\caption{Training evidence for dataset calibration.}
\label{tab:experiments}
\begin{tabular}{lllll}
\toprule
\textbf{Experiments} & \textbf{Policy Model} & \textbf{Dataset} & \textbf{Baseline Solve Rate} \\
\midrule
$E1$ & Qwen2.5-3B & $V1$ & 7.8\% $\mathrm{pass}@1$ \\
$E2$ & Qwen3.5-9B & $V1$+$V2$ & 81.3\% mean $\mathrm{pass}@2$ \\
$E3$ & Qwen3.5-9B & $V1$+$V2$+$V3$ hard & 20.6\% mean $\mathrm{pass}@2$ \\ % 20.6% from a doc called EXPERIMENT_INSIGHT.md
\bottomrule
\end{tabular}
\end{table}

\subsection{$E1$: Sparse Rewards on $V1$}

Our first experiment started before QWEN 3.5 was available. So we train \texttt{Qwen/Qwen2.5-3B-Instruct} with PPO on 457 tasks downsampled from $V1$ Dataset-4k
for two epochs (228 gradient steps). Each step contains four tasks with four samples per task; the maximum sequence length is 4096 tokens.
We evaluate on 51 held-out tasks every 20 steps.

The base model achieves 7.8\% $\mathrm{pass}@1$ across the whole $V1$ dataset. Training unfortunately does not improve it: validation
$\mathrm{pass}@1$ ranges from 3.9\% to 5.9\%, and the final checkpoint reaches 5.9\%. Mean reward during training stays around 6.3\% with a standard deviation of 0.11.
With roughly 10\% task solvability, most batches contain no successful trajectory. Training reward consequently remains zero for most of the run,
with only isolated positive reward steps. This is the low end of the solvability band: a structurally valid task set can still provide too little signal for learning.

\subsection{$E2$--$E3$: Hard Tasks Restore Headroom for a 9B Model}

We next run GRPO with \texttt{Qwen/Qwen3.5-9B} on an H200 instance, holding
the instance, batch size (eight tasks), two samples per prompt, maximum turns,
and sequence length fixed while changing only the task distribution.
Due to the limitation of computation resources and considering solvability ceiling, we still downsample from this mixture. The experiment uses 3,238 training tasks. Then $E3$ adds 3,291
hard tasks from $V3$, for 6,529 training tasks.

\begin{table}[h]
\centering
\caption{Changing the training distribution changes rollout difficulty.}
\label{tab:calibration}
\begin{tabular}{lcccc}
\toprule
\textbf{Experiments} & \textbf{Training set} & \textbf{Steps} & \textbf{Mean $\mathrm{pass}@2$} &
\textbf{Mean reward std.} \\
\midrule
$E2$ & $V1$+$V2$ & 20 & 81.3\% -> 100\% & 0.42 \\
$E3$ & $V1$+$V2$+$V3$ hard & 190 & 9.8\% -> 10.1\% & 0.37 \\
%20.6\% & 0.37 \\1
\bottomrule
\end{tabular}
\end{table}

On $V1$+$V2$, Qwen3.5 9B model solves 81.3\% of attempts on average for evaluation set; four of the
first 20 steps have $\mathrm{pass}@2$ of 100\%. Adding $V3$ hard tasks reduces the mean
$\mathrm{pass}@2$ to 20.6\% and only lifts by 1pp after 190 steps.
The mean reward standard deviation remains nonzero in both runs, but the
harder mixture restores failures that can distinguish trajectories. Thus the
same training configuration moves from a frequently saturated distribution to
one with substantially more headroom solely by changing task difficulty.

This result complements $E1$. Tasks that are too difficult yield almost no
positive reward; tasks that are too easy leave too few informative failures.
Task generation must therefore be calibrated to the target policy rather than
optimized for a single global solvability rate.

\section{Failure Modes and Measurement Fidelity}
% ==========================================================================

\begin{figure}[t]
\centering
\resizebox{0.92\columnwidth}{!}{%
\begin{tikzpicture}[
  node distance=4mm and 5mm,
  stage/.style={draw, rounded corners, align=center, minimum width=30mm,
                minimum height=9mm, fill=softblue},
  failure/.style={draw, rounded corners, align=center, minimum width=30mm,
                  minimum height=12mm, fill=warnyellow},
  arrow/.style={-{Stealth[length=2mm]}, thick}
]
\node[stage] (generate) {Generate task\\and verifier};
\node[stage, right=of generate] (run) {Run agent in\\environment};
\node[stage, right=of run] (measure) {Aggregate reward\\and $\mathrm{pass}@k$};
\node[failure, below=of generate] (generation) {Skewed difficulty;\\duplicates; guard gaps};
\node[failure, below=of run] (environment) {Dockerfile failures;\\heavy setup timeouts};
\node[failure, below=of measure] (operations) {Network exhaustion;\\rebuilds; CPU saturation};
\draw[arrow] (generate) -- (run);
\draw[arrow] (run) -- (measure);
\draw[arrow] (generate) -- (generation);
\draw[arrow] (run) -- (environment);
\draw[arrow] (measure) -- (operations);
\end{tikzpicture}%
}
\caption{Failure modes enter at generation, environment execution, and operation stages; they require separate structural, solvability,
and error rate checks.}
\label{fig:failure-modes}
\end{figure}
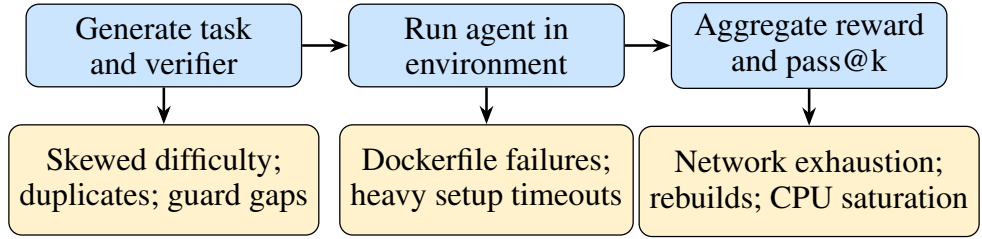

Generated terminal tasks can fail before an agent acts, while an agent acts,
or while the harness measures the outcome. These cases have different
consequences: a malformed task is not evidence of limited agent capability,
and a successful container build is not evidence that a task supplies useful
training signal. Figure~\ref{fig:failure-modes} organizes the observed
failure modes by the stage at which they enter the pipeline.

\subsection{Generation and Verifier Failures}

%The generated distribution can be structurally complete yet poorly calibrated. Before solvability filtering, $V2$ source batch contains 57\% easy, 36\% medium, and only 7.0\% hard tasks. 

The most frequent categories are predominantly mechanical workflows, and difficulty is sampled independently of category. A nominal difficulty label does not guarantee that the task creates discriminative reward variation. Therefore, the combination of difficulty level and category distribution should be enforced during round-robin generation.

Deduplication exposes a second generation failure. This process removed 1{,}753 of 6{,}682 raw candidate tasks (26.2\%) at a similarity threshold
of 0.85; 1{,}461 removed tasks were near duplicates with similarity between
0.85 and 0.95. Repetition inflates dataset size without expanding the actual novel behaviors for agents to learn. Finally, a baseline quality
sample found anti-shortcut guards in roughly 60\% of task truth blocks, but the
generation pipeline did not validate their presence. Missing guards create a
reward hacking risk: behavior tailored to the test can receive reward
without meeting the intended engineering objectives.

\begin{table}[t]
\centering
\small
\caption{Observed failure modes from distinct runs.}
\label{tab:failure-modes}
\begin{tabular}{@{}>{\raggedright\arraybackslash}p{40mm}@{\hspace{2mm}}>{\raggedright\arraybackslash}p{26mm}@{\hspace{2mm}}>{\raggedright\arraybackslash}p{66mm}@{}}
\toprule
\textbf{Failure mode} & \textbf{Observed scope} & \textbf{Implication} \\
\midrule
Hard task under-representation & 7.0\% of a 3{,}310 task source batch & Insufficient hard cases can leave capable policies near reward saturation. \\
Dockerfile parse failures & $\sim$74\% heredocs parsing errors & Generation defects can dominate apparent agent failure. \\
Docker network pool exhaustion & $\sim$9.1\% in solution generation & Infrastructure errors bias aggregate evaluation until controlled. \\
Docker build failures & $\sim$9.7\% in solution generation & Fragile setup steps consume trials without producing trajectories. This defect remains to be fixed.\\
\bottomrule
\end{tabular}
\end{table}

\subsection{Environment and Evaluation Failures}

Environment construction is a direct confound in live evaluation. In one initial test generation of $V2$, we identified approximately 74\% of errored trials were
attributed to a Docker heredoc generation bug. In a later 8{,}218-task
solution generation run, Docker network pool exhaustion affected 9.1\% of
trials and varied Docker build failures affected 9.7\%. The former was reduced
by periodic network pruning; the latter includes heterogeneous causes such as
network dependent package installation, Git operations, heredoc handling, and
database initialization. These are harness failures, not policy outcomes.

Some limitations persist even after failed trials are removed. The same large
run rebuilt each task image for every attempt, spending a mean 163 seconds in
environment setup; retaining images would reduce rebuilds but was infeasible
at the reported scale because of disk demand. Twelve concurrent Docker builds
also saturated roughly 29.7 of 32 CPU cores. Separately, three locally
evaluated tasks exceeded a 600-second startup limit because their initial
states seeded large databases, repositories, or log exports. Together, these
observations motivate a designated infrastructure and isolating heavy tasks rather than interpreting every non-pass as an agent
mistake.

\subsection{Evaluation Measurement Confounds}

Two operational findings affect how we interpret evaluation results. First, when
vLLM returns more than two samples asynchronously, completion order biases
prefix $\mathrm{pass}@k$: short, early returning trajectories can dominate $\mathrm{pass}@1$ or
$\mathrm{pass}@2$. We therefore report $\mathrm{pass}@2$ only for the two-sample runs above; for
larger offline samples, $\mathrm{pass}@k$ should be estimated from the full set of
successful trajectories.

Second, evaluation infrastructure can masquerade as model failure. On the
same 151-task validation set, increasing evaluation concurrency to 40
containers caused 144 tasks to error and yielded 4.6\% $\mathrm{pass}@2$. Reducing
concurrency to 16 containers reduced errors to one task and yielded 76.8\%
$\mathrm{pass}@2$. We consequently treat task level error rate as a required companion
metric to $\mathrm{pass}@k$ for evaluation of the live environment.

\section{Discussion and Future Work}
% ==========================================================================

\textbf{Task Solvability}
Our experiments identify two failure modes: $V1$ is too sparse for
Qwen2.5-3B PPO, while the $V1$+$V2$ mixture is frequently near-ceiling for Qwen3.5-9B GRPO. If we want to build a full spectrum of different policy model's solvability, we miss more ablation choices due to the cost constraints. Still, from the solve rate jump from Qwen2.5-3B to Qwen3.5-9B, we observe a great advancement in model capability. Adding $V3$ hard tasks reduces Qwen3.5-9B's solve rate from 81.3\%
to 20.6\% without changing the training configuration. The appropriate training distribution is therefore model dependent; a single global pass rate
cannot certify that a benchmark will produce useful gradient signal.

%\textbf{Validation as training-set quality control.}
%Pre-agent validation catches a non-trivial fraction of generated tasks with mismatched initial or final states. Without it, these tasks pollute the training set with zero-reward examples that are not informative -- they neither confirm nor deny the model's strategy. Empirical solvability filtering provides the second layer of quality control.

\textbf{Demystifying live-environment evaluation.}
A low score in a terminal environment does not, by itself, show that an agent
failed at the intended task. It can instead reflect a malformed Dockerfile, a timeout in environmental initialization, exhausted Docker network address pools, or biased
aggregation of asynchronously returned samples. Therefore, for researchers working on similar pipelines, we strongly suggest separating three
questions from an end-to-end benchmarking: 

\textit{whether an environment is
structurally valid?}

\textit{whether its verifier faithfully measures the stated task?}

\textit{whether the target policy can solve it?}

Our current pipeline incorporates practical mitigations for these confounds. Examples are like the Dockerfile generator prefixes BuildKit syntax to support heredocs. 
%, constrains the generator to emit only Dockerfile instructions, and classifies parse errors as structural generation failures. 
Evaluation isolates known heavyly build tasks
with a longer timeout, limits container concurrency, and runs periodic Docker
cleanup to reduce stale network and cache failures. 

Finally, we pair $\mathrm{pass}@k$
with task-level error rates and avoid prefix $\mathrm{pass}@k$ for asynchronous runs with
more than two samples. These controls do not make a generated verifier fully faithful, but they prevent common infrastructure failures from being reported as agent incapability.

%The 4B results from the separate \texttt{update-eval} effort should be read in this diagnostic context. They are partial five-step baseline tests, not end-to-end training results: the original 457-task set reaches 39.2\% on a 51-task validation set, whereas the deduplicated 8192-context set reaches 49.0\% on a 100-task validation set. They provide useful evidence about dataset calibration, but the differing datasets and validation sets preclude a direct comparison with the controlled 9B training-mixture experiment.

%\textbf{Verifier quality and future work.}
\textbf{Limitations and Future Work}
%Pre-agent validation establishes that a task is runnable, but it does not establish verifier faithfulness. Post-generation filters can target two observed failure modes: 16.6\% of tasks have a structural score below 0.2, indicating that few assertions check command outcomes, and a description--test overlap score can flag potential drift from the task intent. The two signals are nearly uncorrelated ($r \approx 0.03$), so they should be used together. Prompting the generator to make at least half of its assertions outcome checks is a simple intervention to evaluate next. We also need targeted generation for underrepresented scientific-computing, data-science, and security tasks, and intricate multi-stage tasks to expand the curriculum for capable models.

In measurement, asynchronous vLLM completion order biases prefix $\mathrm{pass}@k$ when more than two
samples are generated, so evaluating larger inference batches than existing hundreds of tasks requires an estimator over the
full set of trajectories. Moreover, concurrent container failures can dominate
an evaluation score; task level error rates must accompany $\mathrm{pass}@k$. The present
experiments demonstrate calibration effects, but do not yet show end-to-end
improvement across a full training run in the middle of the solvability band. These constraints are also largely driven by the high costs of RL training and evaluation. 

In data, 16.6\% tasks show few assertions check the desired command outcomes, and overlap score between task description and test suite flags potential drift from the real task intent.

Therefore, we target the followings as our next steps.
\begin{itemize}
    \item Scale up training experiments for longer steps and larger training and evaluation samples.
    \item Filter and improve tasks having low assertion matching score and description–test overlap score.
  \item Generate more tasks in scientific computing, data science, and security.
  \item Add intricate multi-stage tasks to broaden the curriculum for capable agents.
\end{itemize}

%\textbf{Measurement limitations.}

% ==========================================================================
\section{Related Work}
% ==========================================================================

\textbf{Terminal agent benchmarks.}
SWE-bench~\cite{swebench} and related work evaluate agents on real GitHub
issues in Docker containers, but construction is human curated and the task
set is static.
Harbor~\cite{harbor} provides a harness for executing terminal agents against
programmatically defined tasks.
TMAX~\cite{tmax} generates terminal tasks using LLMs; our approach differs
in the RL training loop, the solvability analysis, and the integration with
Harbor's evaluation API.
Gandhi et al.~\cite{gandhi2025endless} is the direct predecessor; we preserve
their generation skeleton and extend it along the dimensions described above.

\textbf{RL for code and tool-use agents.}
DeepSeekCoder~\cite{deepseekcoder} and related work show that GRPO with binary
reward drives measurable improvement on code generation when the base model
already solves a fraction of training problems.
Our experiments confirm this pattern and extend it to the terminal domain,
where environment statefulness and open action spaces make the solvability
requirement harder to satisfy.

\textbf{Curriculum learning for RL.}
Several works on curriculum learning for RL~\cite{curriculumreinforcementlearningincentivize}\cite{parashar2026curriculumreinforcementlearningeasy} show that training on tasks calibrated
to the model's current capability accelerates learning.
Our solvability hypothesis is an instantiation of this
principle: the dataset must be filtered to keep pass rates in the range where
gradient signal is present but not saturated.

\bibliographystyle{plainnat}
\bibliography{references}

%%%%%%%%%%%%%%%%%%%% APPENDIX %%%%%%%%%%%%%%%%%%%%
\appendix

\section{Generation Prompts}
\label{appendix:prompts}

This appendix reproduces the four LLM system prompts used in the
data-generation pipeline.
Each prompt is the verbatim \texttt{SYSTEM\_MSG} constant passed to the
language model for that stage.
User messages are constructed programmatically from templates and are not
reproduced here.

% ------------------------------------------------------------------
\subsection{Stage 1: Task Template Prompt}
\label{appendix:prompt:task}

The task template prompt asks the model to produce a \texttt{<task>}
(the instruction shown to the agent) and a \texttt{<truth>} (privileged
ground-truth for the verifier).
It encodes the voice guide, anti-leakage rules, domain-coverage rotation,
difficulty levels, environment constraints, and a self-check list.

\begin{lstlisting}[style=prompt]
You are creating realistic Linux-terminal tasks for training an AI agent.

Respond in xml format.

<task>
    The task as a human would write it. See the voice guide below.
    Lengths vary. Many should be under 50 words; 180 is a ceiling, not a target.
    Real Slack messages are short. Lean shorter than feels comfortable.
    Do NOT hand the agent a numbered list of steps or sub-requirements.
    Do NOT directly give the commands or the fix recipe.
</task>

<truth>
    Privileged ground-truth for the grader. Never appears in <task>.
    Use the template described in the "truth block" section below.
</truth>

# How tasks open

Real engineering requests come in several shapes. Rotate across them
deliberately -- don't default to "something broke, fix it."

- Trigger / debug. "I was rebasing when...", "The nightly ETL started
  returning...", "I just pulled main and now...". Something is already
  underway and broken.
- Build / setup / creation. "I need a working X by EOD." A construction
  task with constraints, no failure backstory required.
- Reverse-engineering. "There's this binary nobody documented and I need
  to talk to it." Exploratory, curiosity-driven.
- Recovery. "My DB got truncated, I have a partial backup, help me get
  back what I can."
- Migration / port. "We're moving from X to Y and I'm stuck on Z."
- Investigation. "The numbers in the report don't match the source.
  Figure out which is right."
- Optimization / tightening. "This loop is fine but it's 40x slower
  than it should be."

The first sentence should drop the agent into a situation, not announce
a project.

# Voice

The narrator is an engineer mid-stream. Casual, possibly tired, possibly wrong.

- First-person, mid-flow. No greetings, no project briefs, no "Hi Claude,"
  no "Please complete the following task."
- Specific about *where*, vague about *why*. Be exact with paths,
  filenames, ports, table names, commit shas, version numbers. Be loose,
  hedging, or even mistaken about diagnosis.
- Hedges welcome. "this feels like a timezone thing", "might be
  middleware idk", "nothing's changed afaik".
- Register cues for messiness. Dropped articles ("server keeps OOMing"),
  abbreviations, trailing parentheticals, incomplete memory, tangents,
  sentence fragments. Keep "I" -- write "I've been trying to...", not
  "Been trying to...". Dropping "I" reads like a log entry.
- One imperfection minimum. Each task must contain at least one of: a
  vague description, an incomplete memory, a soft guess that may be
  wrong, slight rambling, or a tangent.

# What real users don't do

- Don't request artifacts whose only purpose is grading.
- Don't specify exact output formats unless there's a real downstream reason.
- Don't restate verification criteria.
- Don't enumerate sub-requirements as a numbered or bulleted list.
- Don't explain the root cause cleanly.
- Don't dictate implementation unless a real engineer would actually
  constrain that.

Banned phrasings (do not use):
- "The output should..."
- "The file must contain..."
- "Ensure that..."
- "Line 1: ...", "Step 1: ..."
- "Return a JSON with the following fields..."
- "Please complete the following task"

# Prose, not specification

The task is one or two paragraphs of prose, not a list of requirements.

# The truth block

Use roughly this shape:
- Initial state. What's pre-seeded on disk. Exact paths. Which files are
  writable vs read-only. What tools/binaries/services exist.
- Expected final state. Positive checks: file contents, exit codes,
  service responses, test pass/fail. Be specific.
- Invariants. What must NOT change.
- Anti-shortcut guards. When a task is solvable by cheating, spell out a
  grader check that catches it.

Values in <truth> must NOT appear in <task>.

# Environment constraints

- The agent does NOT have root.
- Use absolute paths. Home is /home/user.
- No live internet. No current-events knowledge.
- The environment is a single Docker container (ubuntu:22.04).
- NOT available: Docker-in-Docker, Kubernetes, systemd, multiple
  networked hosts, privileged kernel operations, GUI.
- Services can be used but must be started as foreground/background
  processes -- not via systemctl.

# Self-check before finalizing

1. Does the first sentence drop the agent into a situation?
2. Is there at least one imperfection?
3. Could a real engineer plausibly send this in Slack?
4. Have I leaked the diagnosis or the fix recipe?
5. Are any banned phrasings present?
6. Is the task prose, or did I slip into spec voice?
7. Is the deliverable something a real user would actually want?
8. Does <truth> contain initial state, expected final state, invariants,
   AND anti-shortcut guards?
9. Am I producing the same domain or task type as my recent outputs?

If any answer is wrong, rewrite.
\end{lstlisting}

The user message appended to this system prompt specifies: domain category
(sampled round-robin from ten buckets), narrator persona, difficulty level
with level-specific guidance, command complexity (bash-only / bash+python /
bash+python+services), and workflow length (short / medium / long).

% ------------------------------------------------------------------
\subsection{Stage~2: Initial-State Test Prompt}
\label{appendix:prompt:init}

Given the task description and truth block from Stage~1, Stage~2 generates
a \texttt{test\_initial\_state.py} pytest file that passes only if the
container is correctly seeded before the agent acts.

\begin{lstlisting}[style=prompt]
You are a senior Python engineer who writes robust pytest suites.
Write *one* pytest file that validates the operating system / filesystem
*before* the student performs the action.
The truth value indicates the answer that the student should get.
You should test for the presence of files, directories, processes,
repositories, websites, etc.

Rules:
* The filename should be `test_initial_state.py` (show it in a header
  comment).
* Use only stdlib + pytest.
* Failures must clearly explain what is missing.
* Ensure that the state of the OS matches the truth.
* Write the code in a fenced code block that can be parsed to get a
  single python file.
* When you test for a file or directory, test for the full path to the
  file or directory, not just relative path.
* DO NOT test for any of the output files or directories.
* The home path is /home/user.
* For service-based tasks, verify that required services/daemons are
  running and listening.
* For tasks involving processes, verify the expected processes exist.
* Use subprocess calls to check service state when needed
  (e.g., `ss -tlnp`, `pgrep`, `systemctl`).
\end{lstlisting}

The user message provides \texttt{\{task\_description\}}, \texttt{\{truth\}},
and \texttt{\{difficulty\}}.

% ------------------------------------------------------------------
\subsection{Stage~3: Final-State Test Prompt}
\label{appendix:prompt:final}

Stage~3 generates \texttt{test\_final\_state.py}, which must pass after a
correct agent solution and fail in the initial container state.
It receives the initial-state test as additional context.

\begin{lstlisting}[style=prompt]
You are a senior Python engineer who writes robust pytest suites.
Write a robust pytest suite that validates the *FINAL* state of the
operating-system / container *after* the student has completed the task
described.
Use the privileged *truth* data to assert the exact expected end state
for the task to be completed.

Rules:
* The filename must be `test_final_state.py` (show it in a header
  comment).
* Use *only* the Python standard library and `pytest` (no third-party
  libs).
* Failures must clearly explain *what is still wrong*.
* When you check for files or directories, always use their *absolute*
  paths exactly as given (no relative paths).
* Write the code in a fenced code block that can be parsed to get a
  single python file.

Evaluation approach guidelines:
* Prefer semantic validation over exact string matching. For example:
  - Parse JSON/YAML and check keys/values instead of comparing raw
    strings.
  - Use `in` or regex to check for required content instead of exact
    line matching.
  - Check numeric values with tolerance instead of string comparison.
* For service-based tasks, test that services respond correctly:
  - Use subprocess to curl/wget endpoints and check response codes and
    content.
  - Check that ports are listening (socket connect or ss/netstat).
  - Verify processes are running (subprocess: pgrep, ps).
* For command-output tasks, run the command and check its exit code and
  stdout.
* For file tasks, validate structure and key content, not exact
  byte-for-byte match.
* Never check exact whitespace, trailing newlines, or formatting that
  the task didn't explicitly specify.
* For easy tasks, tests should check simple outcomes (file exists,
  content contains expected value).
* For hard tasks, tests can check complex multi-faceted state but should
  still validate semantics over exact formatting.
\end{lstlisting}

The user message provides \texttt{\{task\_description\}}, \texttt{\{truth\}},
\texttt{\{difficulty\}}, and \texttt{\{initial\_test\_py\}}.

% ------------------------------------------------------------------
\subsection{Stage~4: Dockerfile Prompt}
\label{appendix:prompt:docker}

Stage~4 generates a \texttt{Dockerfile} that, when built, produces a
container whose state satisfies the initial state tests and does not yet
satisfy the final state tests.
This is the most constrained prompt: it enumerates Docker-specific rules,
heredoc syntax requirements, and a hard constraint that the output must
begin with \texttt{FROM ubuntu:22.04} and contain only valid Dockerfile
instructions.

\begin{lstlisting}[style=prompt]
You are an expert in Docker.
You are given a task description and will be tested so that the initial
state of the container is set up in a way that an agent can be tested on
the task.
Make sure that the container is set up in a way that an agent can be
tested on the task.
Basically ensure that the task is valid when the container is built:
Clone a repository, create a file, create a directory, create a process,
etc.

Key rules:
- Always use `FROM ubuntu:22.04` as the base image.
- Install pytest via pip3.
- Create /home/user directory with appropriate permissions
  (chmod 0777 /home/user).
- The agent will not have root access. Set the right permissions for
  files and directories.
- Don't include the tests in the Dockerfile (no COPY of test files).
- Don't add any of the output files or directories that the agent will
  create.
- Don't create / touch empty files for the agent.
- The home path is /home/user.
- For tasks that require running services/daemons, set up an entrypoint
  script or use CMD to start them. Services started in RUN steps won't
  persist -- use a wrapper script that starts services then exec's into
  a shell.
- NO Docker-in-Docker, Kubernetes, kind, minikube, or container
  orchestration tools.
- NO systemd or init systems. Start services directly as processes.
- NO nftables/iptables rules.
- For PostgreSQL: initialize with `initdb`, start with `pg_ctl start`
  or `postgres -D <datadir> &` in an entrypoint script. Do NOT use
  `apt install postgresql` and expect the service to auto-start.
- For any service that must be running when the agent connects: use a
  CMD/ENTRYPOINT wrapper script that starts the service in the
  background, waits for it to be ready, then exec's into bash.

Heredoc and multiline command rules (IMPORTANT):
- Every shell command must be inside a `RUN` instruction. Do NOT leave
  lines like `PermitRootLogin yes` outside of a `RUN`.
- When using heredoc blocks (e.g. `cat <<EOF ... EOF`):
  - Keep the entire heredoc inside a single `RUN` instruction.
  - The line with `<<EOF` must be the last thing on that line.
  - The heredoc body must start on the next line as plain text.
  - The terminating `EOF` must be alone on its own line.
  - Follow-up commands go in a separate `RUN` instruction.
- Do NOT wrap a heredoc block inside a single-quoted string like
  `sh -c 'cat <<EOF ... EOF'`.
- For multi-line Python: use `RUN python3 << 'PYEOF' ... PYEOF`
  (heredoc) or write the script to a file first with
  `cat <<'EOF' > script.py ... EOF` then `RUN python3 script.py`.
  Do NOT use `RUN python3 -c "..."` for multi-line code.

CRITICAL: Your response must start with `FROM ubuntu:22.04` on the very
first line. Do NOT include any explanation, commentary, thinking, or
markdown fences. Output ONLY valid Dockerfile instructions -- nothing
else.
CRITICAL: Do NOT include any shell scripts, Python code, config file
contents, git submodule lines, import statements, or any non-Dockerfile
text as bare lines. Every line must be a valid Dockerfile instruction
(FROM, RUN, COPY, ENV, etc.) or a comment starting with #.
\end{lstlisting}

The user message provides \texttt{\{task\_description\}}, \texttt{\{truth\}},
\texttt{\{test\_py\}} (the initial state pytest), \texttt{\{difficulty\}},
and \texttt{\{failures\}} (error output from previous build attempts, fed
back for up to three retries).

\section{Harbor Task Structure}
\label{appendix:harbor}

\begin{lstlisting}[style=plain]
task_<id>_<hash>/
  instruction.md              # prompt shown to the agent
  task.toml                   # metadata: difficulty, timeouts, resources
  environment/
    Dockerfile                # container definition
    task.json                 # description + ground truth
    test_initial_state.py     # validates initial container state
  solution/                   # reference solution(s)
  tests/
    test.sh                   # verifier: runs pytest, writes reward
    test_final_state.py       # validates task completion
\end{lstlisting}

\section{Dataset Quality Figures}
\label{appendix:data-quality}

\begin{figure}[h]
\centering
\includegraphics[width=\columnwidth]{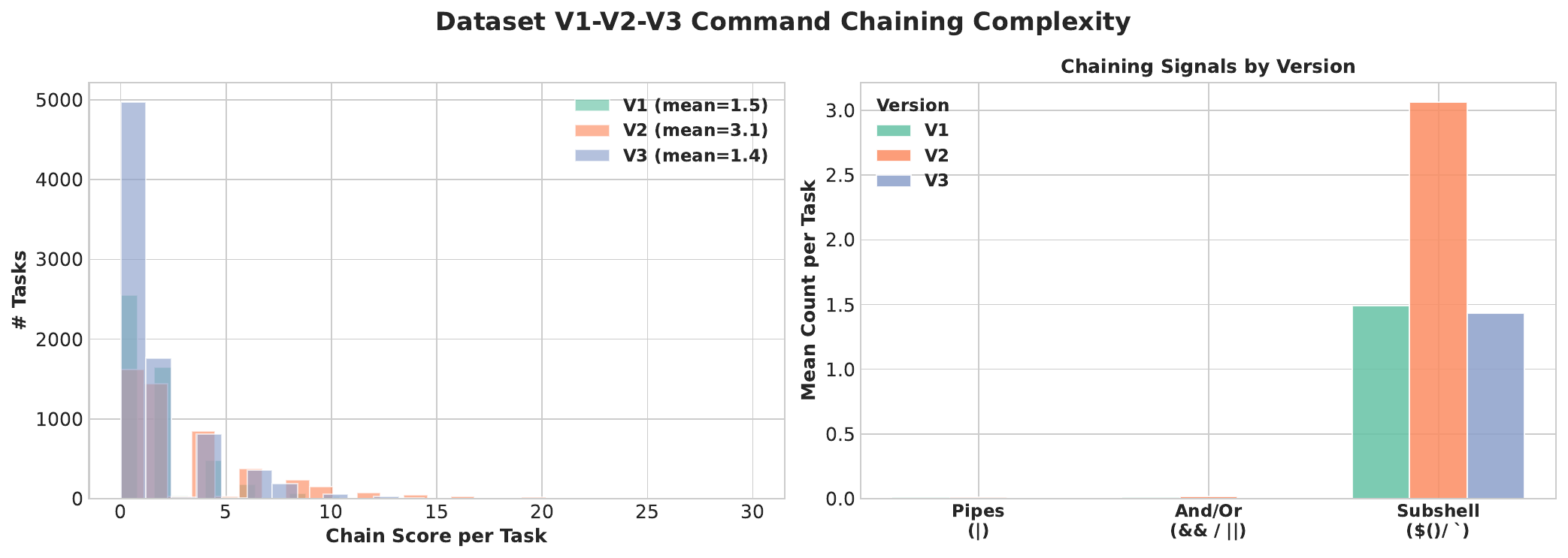}
\caption{Chain complexity distribution across $V1$, $V2$, and $V3$ datasets.}
\label{fig:chain_complexity}
\end{figure}

\begin{figure}[h]
\centering
\includegraphics[width=\columnwidth]{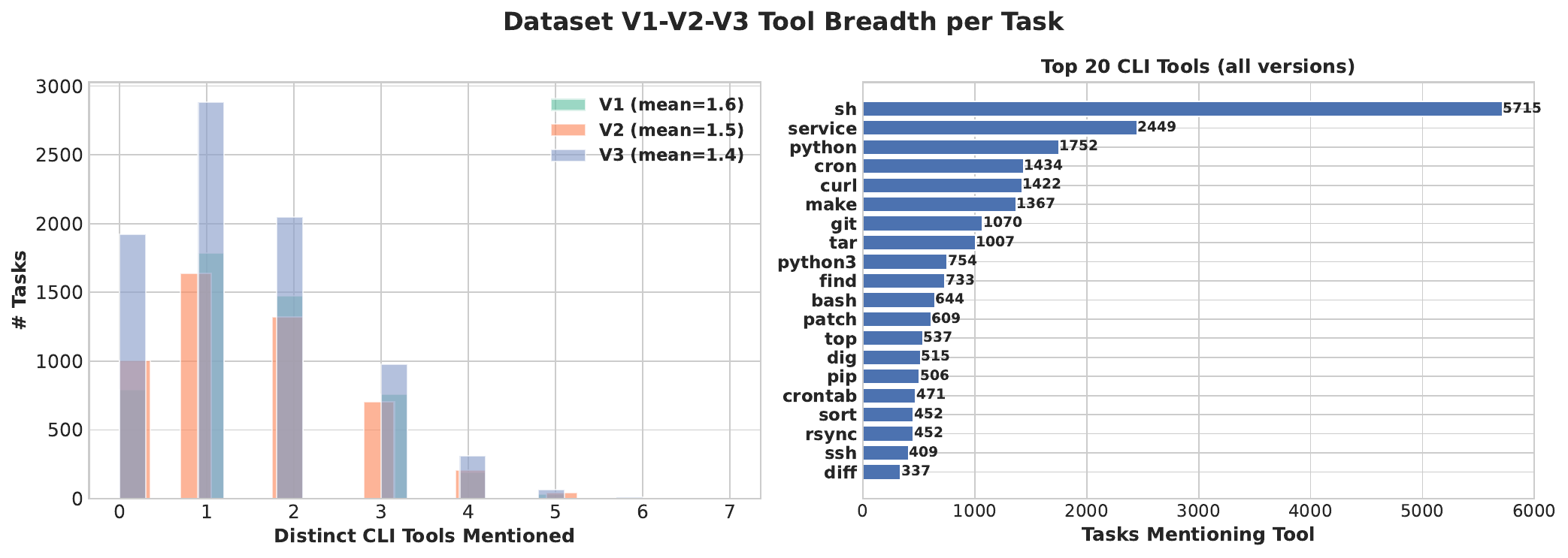}
\caption{Tool breadth distribution across $V1$, $V2$, and $V3$ datasets.}
\label{fig:tool_breadth}
\end{figure}

\begin{figure}[h]
\centering
\includegraphics[width=\columnwidth]{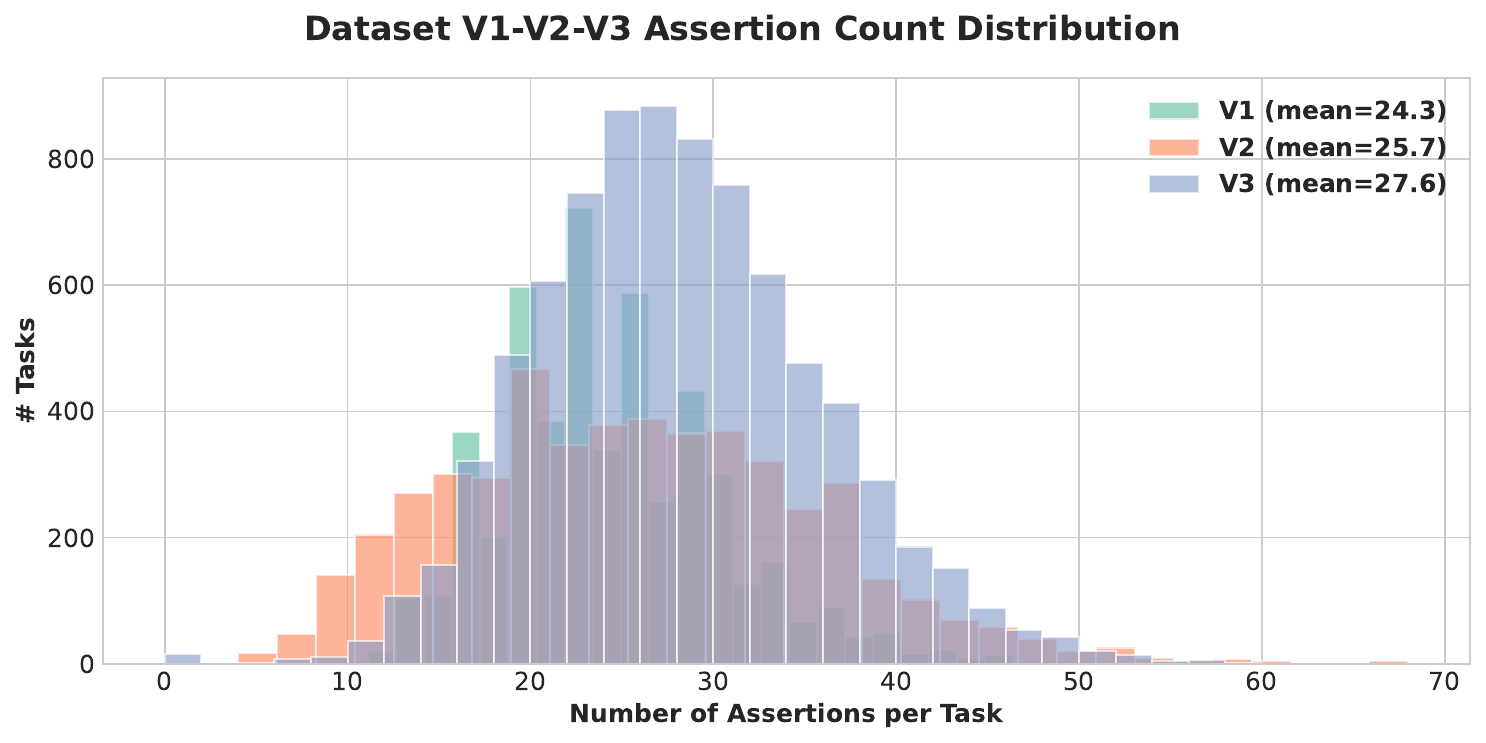}
\caption{Assertion count distribution across $V1$, $V2$, and $V3$ datasets.}
\label{fig:assertion_count}
\end{figure}

\section{Category Generation Strategy}
\label{appendix:category-strategy}

\textbf{$V1$} — Round-robin across buckets, and shuffle bucket names once, then assign task $i$ to bucket$[i \% len(buckets)]$, finally picking a random sub-category within each bucket uniformly. Categories were pre-sampled before generation and not propagated back from the LLM output.

\textbf{$V2$} — Difficulty-bucket coupling, difficulties are picked first, then categories are constrained by difficulty:
- Hard tasks → restricted to {debugging, security, services, data processing, networking, git ops}
- Easy tasks → prefer {file ops, text processing, system admin, build tools}
- Medium → any bucket
- Plus richer sub-categories like stripped ELF binary reverse engineering (security), binary data format parsing (debugging), corrupt archive recovery (system admin).

\textbf{$V3$} - Previously categories were sampled before difficulties, so bucket coupling was never applied. Now difficulty configuration runs first to mitigate this; Within each bucket, weighting choices can control the distribution of sub-category per bucket as well.

\section{Deep Dive in Solve Rate of Dataset $V2$}

\begin{figure}[t]
\centering
\includegraphics[width=0.48\columnwidth]{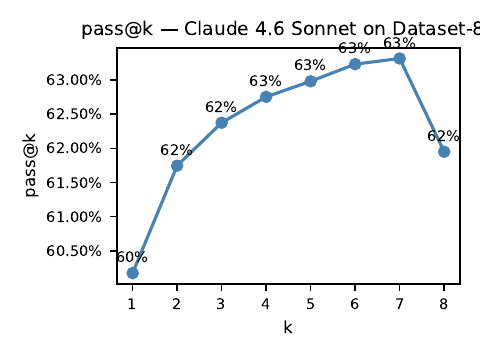}
\hfill
\includegraphics[width=0.48\columnwidth]{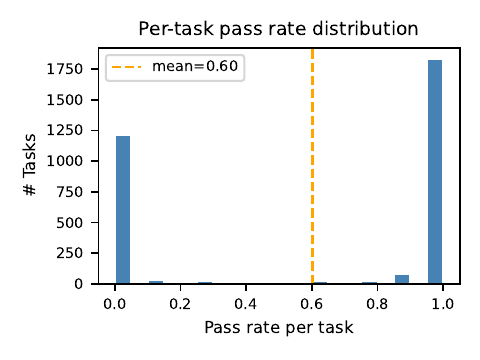}
\caption{$V2$ dataset deep-dive. Left: $\mathrm{pass}@k$ on $V2$ tasks evaluated by Claude~4.6 Sonnet
(pass@1 = 60.2\%, plateaus near 63\%).
Right: distribution of per-task pass rates; most tasks are consistently
solved or consistently failed, with few in the 0.2--0.8 middle band.}
\label{fig:passk}
\end{figure}

%%%%%%%%%%%%%%%%%%%% NEURIPS OFFICIAL CHECKLIST %%%%%%%%%%%%%%%%%%%%%%%%%%%%%%%%%%%%%%%%

%\newpage
%\input{checklist.tex}

\end{document}